\documentclass[10pt,conference,letterpaper]{IEEEtran}

\usepackage{cite}
\usepackage{amsmath,amssymb}
\usepackage{graphicx}
\usepackage{booktabs}
\usepackage[table]{xcolor}
\usepackage{algorithm}
\usepackage{algorithmic}
\usepackage{balance}

\newcommand{\sys}{DigenRL}

\begin{document}

\title{\sys{}: Accelerating Disaggregated RL for Visual Generative LLMs with
Diffusion-Based Parallelism and Trainer-Assisted Generation}

\author{
\IEEEauthorblockN{Sijie Wang\IEEEauthorrefmark{1}, Zhengyu Qing\IEEEauthorrefmark{2},
Tan Zhiqiang\IEEEauthorrefmark{3}, and Yiming Yin\IEEEauthorrefmark{4}\\
Yeqing Zhang\IEEEauthorrefmark{5}, Yaoyuan Wang\IEEEauthorrefmark{5},
Qiang Wang\IEEEauthorrefmark{2}, Xiaowen Chu\IEEEauthorrefmark{6},
and Shaohuai Shi\IEEEauthorrefmark{2}}
\IEEEauthorblockA{\IEEEauthorrefmark{1}Harbin Institute of Technology (Shenzhen), China}
\IEEEauthorblockA{\IEEEauthorrefmark{2}Harbin Institute of Technology, Shenzhen, China}
\IEEEauthorblockA{\IEEEauthorrefmark{3}Harbin Institute of Technology, China}
\IEEEauthorblockA{\IEEEauthorrefmark{4}The Hong Kong University of Science and Technology (Guangzhou), China}
\IEEEauthorblockA{\IEEEauthorrefmark{5}Huawei Technologies Ltd., China}
\IEEEauthorblockA{\IEEEauthorrefmark{6}The Hong Kong University of Science and Technology (Guangzhou)\\
and The Hong Kong University of Science and Technology, Hong Kong}
}

\maketitle

% !TEX root = ../main.tex
\begin{abstract}
Reinforcement learning (RL) has become a dominant post-training paradigm, driving the emergence of high-performance RL systems such as veRL for autoregressive large language models (LLMs). 
In parallel, diffusion-oriented RL algorithms (e.g., DanceGRPO and FlowGRPO) have rapidly expanded the scope of RL from language reasoning to diffusion-based visual and flow-based generation. 
However, efficient RL systems for diffusion generative LLMs remain underexplored. Existing implementations (e.g., veRL-Omni)
still rely on colocated execution, which simplifies synchronization but couples rollout and training resources, limits heterogeneous deployment and constrains independent scaling. 
    
To this end, we introduce \sys{}, a disaggregated RL framework for diffusion-based generative LLMs that supports flexible resource allocation, accommodates heterogeneous GPUs, and facilitates efficient task scheduling. 
To maximally reduce the execution bubbles in the disaggregated architecture, we propose 1) a generation-axis pipeline (GAP) and time-step parallelism (TSP) in the diffusion architecture to enable finer-grained pipelining between rollout and training, 2) an elastic trainer-assisted generation (TAG) approach to enable the trainer GPU resources to dynamically assist in executing rollout generations, and 3) a tightly one-step constrained asynchronous strategy to further utilize the tail bubble in the pipeline. 
Extensive experiments are conducted on three hardware testbeds with 16--32 GPUs using HunyuanVideo-13B, Wan2.1-14B, FLUX.1-12B, QwenImage-20B generative models. Experimental results show that \sys{} achieves 1.56--2.10$\times$ throughput improvements over veRL-Omni.
\end{abstract}

\begin{IEEEkeywords}
diffusion reinforcement learning, disaggregated post-training,
pipeline scheduling, resource scheduling
\end{IEEEkeywords}

% !TEX root = ../main.tex
\section{Introduction}
\label{sec:introduction}

Reinforcement learning (RL) has re-emerged as one of the central paradigms for post-training large language models (LLMs)~\cite{ouyang22,guo2025deepseek}. 
In preference alignment, complex reasoning, and agentic
training, RL typically boosts final model quality using proximal policy optimization (PPO)~\cite{ppo17}, group relative policy optimization (GRPO)~\cite{deepseek2025r1}, or its variants (e.g., DanceGRPO~\cite{dancegrpo25}, FlowGRPO~\cite{flowgrpo25}).
The basic workflow of RL for LLMs proceeds iteratively through: 1) sampling with the current policy model (generator or rollout), 2) evaluating the sampled outputs (reward model or critic), and 3) updating the policy based on these samples (actor update). The latter two steps are often grouped into a single trainer component, which computes rewards and losses to adjust the policy, after which the updated policy is synchronized with the generator. As a result, the generator and trainer share strong data dependencies, demand significant computation, and in practice often become the primary performance bottleneck.
This has motivated the development of high-performance
RL systems such as DeepSpeed-Chat~\cite{yao2023deepspeed}, veRL~\cite{hybridflow24}, AReaL~\cite{areal25}, StreamRL~\cite{streamrl25}, OpenRLHF~\cite{openrlhf24}, LlamaRL~\cite{llamarl25}, ROLL~\cite{roll25}, etc. 
These systems illustrate that the training throughput of RL post-training is influenced not only by the choice of learning algorithms (e.g., PPO~\cite{ppo17}, GRPO~\cite{shao2024deepseekmath}, etc.), but also strongly by the way resources are organized and tasks are scheduled within the system. 

Recent RL systems (e.g., AReaL~\cite{areal25}, StreamRL~\cite{streamrl25}, Laminar~\cite{sheng2026laminar}, OrchestrRL~\cite{tan2026orchestrrl}, and WEAVE~\cite{weave26}) tend to exploit a disaggregated architecture due to its flexibility for resource allocation, asynchronous rollouts, and long-tail generation repacking, which are particularly effective for autoregressive LLM post-training (AR-RL). 
However, for the RL post-training of diffusion-based visual generative models (Diffusion-RL)~\cite{dancegrpo25, flowgrpo25}, current optimized systems (e.g., Long-RL~\cite{longrl25}, GenRL~\cite{genrl25} and veRL-Omni~\cite{verlomni26}) still adopt a colocated architecture. This may be because 1) rollout workers have relatively uniform workloads, and 2) rollout and training share the same backend (e.g., PyTorch-FSDP~\cite{zhao2023pytorch}), avoiding efficiency issues.
Yet, we argue that Diffusion-RL still needs a disaggregated architecture to enable more fine-grained parallelism and task scheduling, thereby making better use of hardware resources and eventually increasing overall system throughput.

\begin{figure}[!t]
  \centering
  \includegraphics[width=\columnwidth]{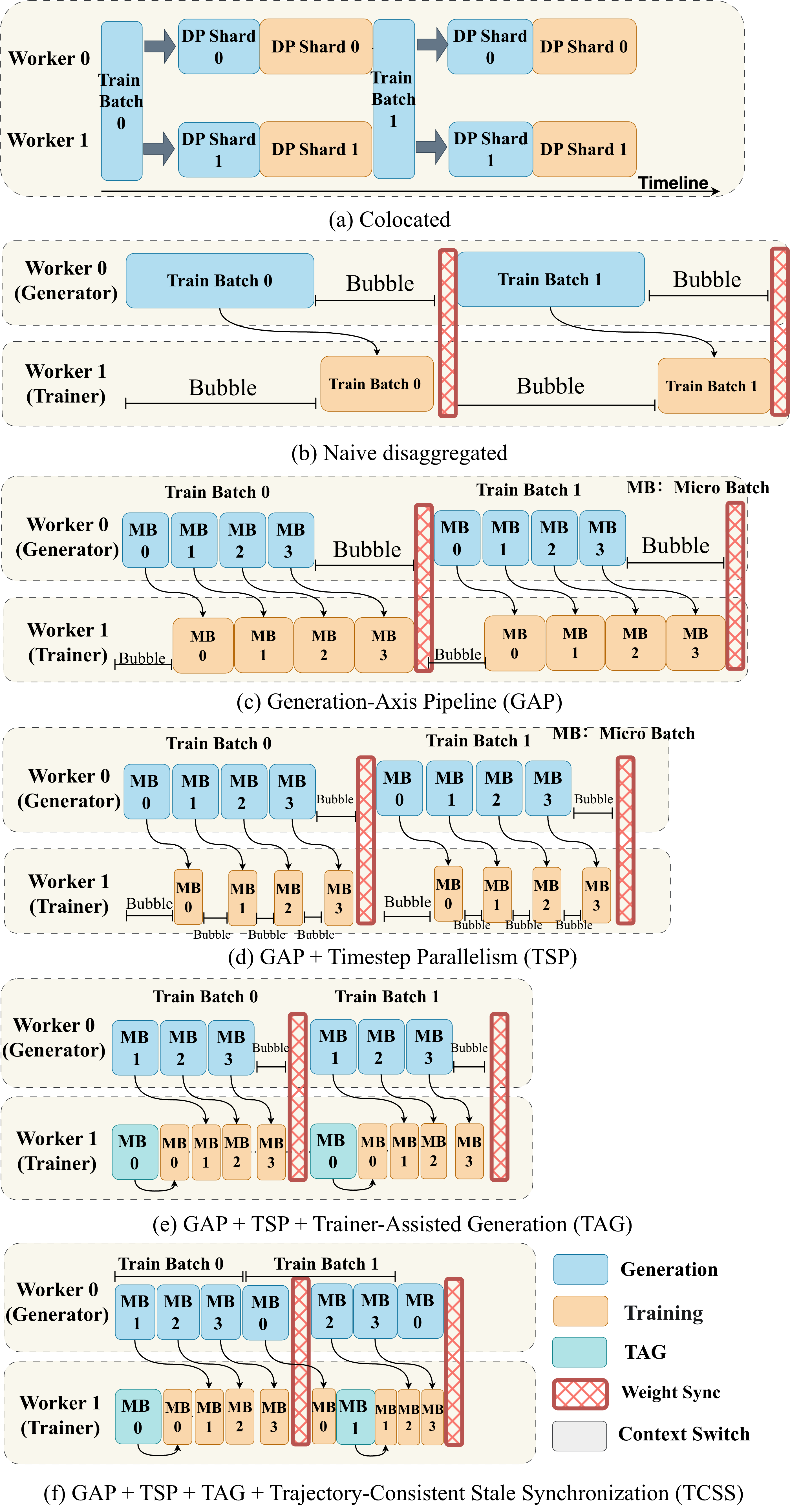}
  \caption{Comparison of colocated and disaggregated Diffusion-RL execution.
  The short reward stage is omitted. (a) Colocated execution. (b) Naive
  disaggregated execution. (c) Generation-axis pipelining (GAP). (d) GAP with
  timestep parallelism (TSP). (e) GAP and TSP with trainer-assisted generation
  (TAG). (f) The complete design with trajectory-consistent stale
  synchronization (TCSS).}
  \label{fig:4methods_com}
\end{figure}

However, as illustrated in Figure~\ref{fig:4methods_com}(b), the disaggregated architecture introduces additional bubbles relative to the collocated architecture in Figure~\ref{fig:4methods_com}(a). Therefore, to enable disaggregated Diffusion-RL to fully exploit GPU resources during training, it is necessary to design fine-grained pipeline scheduling strategies that minimize these bubbles as much as possible.

In this paper, we introduce \sys{}, a disaggregated Diffusion-RL system for improving post-training throughput. To achieve this goal, we propose the following novel techniques. First, we design a diffusion-based parallelism (\S\ref{sec:diffusion-parallelism}) according to the diffusion inference and training characteristics with 1) \emph{generation-axis pipeline (GAP)} execution in the generator and 2) \emph{time-step parallelism (TSP)} in the trainer. GAP and TSP together support finer-grained generation and training task pipelining and reduce the bubble time. Second, to further reduce the bubbles when the trainer waits for the generator, we propose \emph{elastic trainer-assisted
generation (TAG)} to allow the trainer to execute part of the micro-batch
generation workloads when it would otherwise be idle, as shown in Figure~\ref{fig:4methods_com}(d). Third, to better exploit the final micro-batch bubble, which in theory cannot be removed in synchronous Diffusion-RL, we propose a \emph{trajectory-consistent stale synchronization (TCSS)} scheme. TCSS allows the generator to progress by at most one policy version while ensuring that each denoising trajectory is produced entirely under a single, fixed policy snapshot. The feature comparison of our \sys{} with existing RL systems is shown in Table~\ref{tab:digenrl_comparison}. We conduct extensive experiments compared with state-of-the-art Diffusion-RL system veRL-Omni~\cite{verlomni26} using various visual generative models (HunyuanVideo-13B~\cite{hunyuanvideo24}, Wan2.1-14B~\cite{wan25}, FLUX.1-12B~\cite{flux24}, QwenImage-20B~\cite{qwenimage25}) on two homogeneous 32-GPU testbeds and one heterogeneous 16-GPU testbed. Experimental results (\S\ref{sec:evaluation}) show that our \sys{} achieves 1.56--2.10$\times$ speedups over veRL-Omni.

% \begin{table}[t]
%     % \vspace{0.7\baselineskip}
%     \centering
%     \scriptsize
%     \setlength{\tabcolsep}{2pt}
%     \renewcommand{\arraystretch}{0.72}
%     \caption{Comparison of RL frameworks for diffusion models.}
%     \begin{tabular}{@{}M{0.17\columnwidth}M{0.18\columnwidth}M{0.18\columnwidth}M{0.16\columnwidth}M{0.22\columnwidth}@{}}
%         \toprule
%         & \textbf{VeRL/AReaL} & \textbf{GenRL} & \textbf{VeRL-Omni} & \textbf{\sys{}} \\
%         \midrule
%         \textbf{\makecell{Diffusion\\Support}} &
%         \xmark &
%         \makecell{Wan2.1/2.2\\HunyuanImage} &
%         \makecell{Qwen-Image} &
%         \makecell{Wan2.1/2.2\\HunyuanVideo\\Flux\\Qwen-Image} \\
%         \midrule
%         \textbf{\makecell{Algorithm\\Support}} &
%         \makecell{PPO, GRPO\\DAPO, DPO\\ReMax, \dots} &
%         \makecell{FlowGRPO\\MixGRPO, CPS\\LongCat-Video} &
%         \makecell{FlowGRPO} &
%         \makecell{DanceGRPO\\FlowGRPO} \\
%         \midrule
%         \textbf{\makecell{Disaggregate\\Support}} &
%         \makecell{Pipeline overlap\\Partial rollout} &
%         \xmark &
%         \xmark &
%         \makecell{Generation-Axis\\Pipeline (GAP),\\Trainer-Assisted\\Generation (TAG)} \\
%         \midrule
%         \textbf{\makecell{Async.\\Support}} &
%         \makecell{One-Step-off-\\Policy\\Fully async} &
%         \xmark &
%         \xmark &
%         \makecell{Trajectory-\\Consistent\\Stale Sync\\(TCSS)} \\
%         \bottomrule
%     \end{tabular}
%     \label{tab:digenrl_comparison}
%     % \vspace{-1.8\baselineskip}
% \end{table}

\begin{table}[t]
\centering
\small
    \setlength{\tabcolsep}{2pt}
    \renewcommand{\arraystretch}{0.72}
\caption{Comparison of RL frameworks for diffusion models.}
\label{tab:digenrl_comparison}

\rowcolors{2}{gray!8}{white}

\begin{tabular}{lcccc}
\toprule
\textbf{Feature}
& \textbf{veRL}
& \textbf{GenRL}
& \textbf{veRL-Omni}
& \textbf{\sys{}} \\
\midrule

Diffusion-RL
& $\times$
& $\checkmark$
& $\checkmark$
& $\checkmark$ \\

Multi-model Support
& $\times$
& Partial
& Partial
& $\checkmark$ \\

Algorithm Extensibility
& $\checkmark$
& $\checkmark$
& Limited
& $\checkmark$ \\

Disaggregated Training
& Partial
& $\times$
& $\times$
& $\checkmark$ \\

Asynchronous RL
& $\checkmark$
& $\times$
& $\times$
& $\checkmark$ \\

\bottomrule
\end{tabular}
\end{table}

% !TEX root = ../main.tex
\section{Background and Motivation}
\label{sec:background}

\subsection{Diffusion-RL Workflow}
\label{sec:diffusionrl-workflow}

Existing Diffusion-RL algorithms such as DanceGRPO~\cite{dancegrpo25} and
Flow-GRPO~\cite{flowgrpo25} are typically executed in a colocated fashion:
the same worker group, or tightly coupled device pool, is responsible for
both sample generation and trainer-side policy updates. Given a training
batch of size $B$, the generator samples each prompt $M$ times; $b$ indexes
a prompt and $m$ indexes one of its generated samples.
The generator uses the current policy $\pi_k$ to produce a denoising
trajectory~\cite{ddpm20,guidedDiffusion21,flowMatching23}
\begin{align}
  \tau_{b,m}
    &= \{(x_{b,m,t},x_{b,m,t-1},\ell^{\mathrm{old}}_{b,m,t})\}_{t=1}^{T},
    \label{eq:denoising-trajectory}\\
  x_{b,m,t-1}
    &\sim \pi_k(\cdot\mid x_{b,m,t},c_b).
    \label{eq:denoising-transition}
\end{align}
where $x_{b,m,t}$ is the latent at denoising step $t$, $c_b$ is the condition
or prompt, and
\begin{equation}
  \label{eq:rollout-log-probability}
  \ell^{\mathrm{old}}_{b,m,t}
  = \log\pi_k(x_{b,m,t-1}\mid x_{b,m,t},c_b)
\end{equation}
is the step-level log-probability recorded during rollout. 
After rollout, terminal rewards are computed and normalized within each prompt group to obtain
advantages $A_{b,m}$.

The trainer recomputes the step-level log-probability $\ell_{b,m,t}(\theta)$ under the current
policy $\theta$ and forms an importance ratio
\begin{equation}
  \label{eq:importance-ratio}
  \rho_{b,m,t}(\theta)
  = \exp\!\Big(
    \ell_{b,m,t}(\theta)
    - \ell^{\mathrm{old}}_{b,m,t}
  \Big).
\end{equation}

Policy optimization is typically performed only on a selected subset of
denoising steps to reduce training cost~\cite{dancegrpo25,flowgrpo25}. Let
$\mathcal{S}_{b,m}\subseteq\{1,\ldots,T\}$ denote the selected training
timesteps for the $m$-th sample generated for prompt $b$. A PPO/GRPO-style clipped objective can then be
written as
\begin{equation}
  \mathcal{L}_b
    =-
      \sum_{m=1}^{M}\sum_{t\in\mathcal{S}_{b,m}}\mathcal{L}_{b,m,t}.
  \label{eq:per-prompt-policy-loss}
\end{equation}
\begin{equation}
  \begin{aligned}
    \mathcal{L}_{b,m,t}
      &=\min\!\Big(
        \rho_{b,m,t}(\theta)A_{b,m},\,
        \\
      &\qquad
        \operatorname{clip}\!\big(
          \rho_{b,m,t}(\theta),1-\epsilon,1+\epsilon
        \big)A_{b,m}
      \Big).
  \end{aligned}
  \label{eq:clipped-policy-objective}
\end{equation}
Let $Z=\sum_{b=1}^{B}\sum_{m=1}^{M}|\mathcal{S}_{b,m}|$ denote the total number of selected
training timesteps in the batch. The normalized policy-gradient loss and its
gradient are
\begin{align}
  \mathcal{L}_{\mathrm{pg}}
    &=-\frac{1}{Z}\sum_{b=1}^{B}\sum_{m=1}^{M}
      \sum_{t\in\mathcal{S}_{b,m}}\mathcal{L}_{b,m,t},
      \label{eq:normalized-policy-loss}\\
  \nabla_\theta\mathcal{L}_{\mathrm{pg}}
    &=-\frac{1}{Z}\sum_{b=1}^{B}\sum_{m=1}^{M}
      \sum_{t\in\mathcal{S}_{b,m}}
      \nabla_\theta\mathcal{L}_{b,m,t}.
      \label{eq:normalized-policy-gradient}
\end{align}

\subsection{Why Is a Disaggregated Diffusion-RL Framework Needed?}
\label{sec:disaggregated-diffusion-rl}

\begin{figure}[t]
  \centering
  \includegraphics[width=\columnwidth]{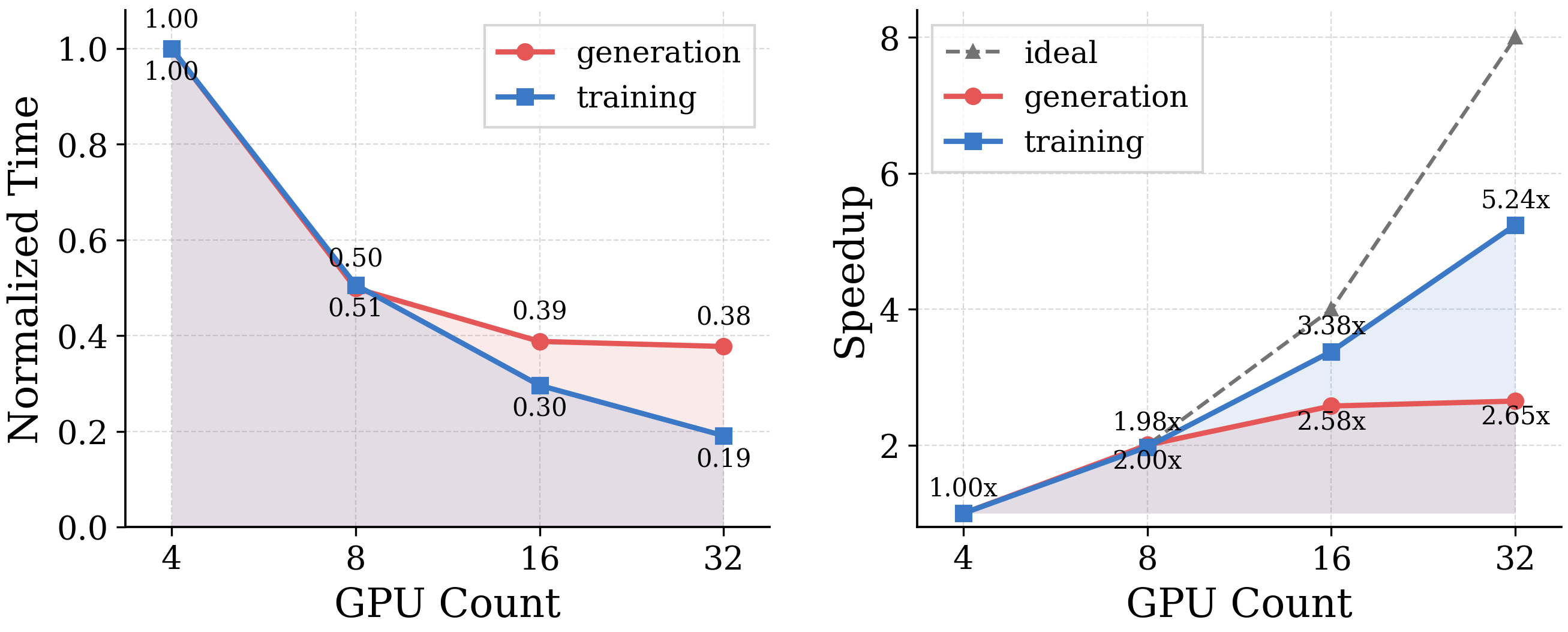}
  \caption{Scalability comparison between generation and training.}
  \label{fig:gpu-stage-scalability}
\end{figure}

Compared with the colocated architecture, in which the generator and trainer
jointly use the entire GPU pool (Figure~\ref{fig:4methods_com}(a)), a
disaggregated architecture splits the GPUs into two non-overlapping sets: a
generator group and a trainer group. The generator and trainer then alternate 
their workloads, as illustrated in
Figure~\ref{fig:4methods_com}(b). During each RL iteration, the generator
produces a micro-batch of training data, which the trainer consumes to update
the model weights; the updated weights are subsequently synchronized back to
the generator. Disaggregated Diffusion-RL is particularly useful when the
generator and trainer have different workloads or when the GPU cluster is
heterogeneous~\cite{sheng2026laminar,streamrl25,rlboost26}.

In Diffusion-RL, the generator advances sequentially along the denoising
chain, whereas trainer-side optimization can parallelize computation across
multiple selected timesteps. This asymmetry can easily create a workload
imbalance between the two stages. As shown in
Figure~\ref{fig:gpu-stage-scalability}, when training FLUX.1-12B on a 32-GPU
cluster, training maintains good scalability as the number of GPUs increases
because it is compute-bound. In contrast, generation must process denoising
steps sequentially and therefore often fails to utilize additional compute
resources effectively.

\subsection{Motivation}
\label{sec:motivation}

As shown in Figure~\ref{fig:4methods_com}(b), naively disaggregating generation
and training in Diffusion-RL produces substantial bubbles.

\textbf{First, naive disaggregation causes mutual waiting.}
The trainer cannot compute losses and gradients until it receives rollout
outputs from the generator, while the generator must wait for the latest
weights from the trainer before starting the next round of generation. These
strict dependencies leave substantial GPU resources idle. Pipelining can
overlap the two stages and mitigate this waste, but the small batch sizes
typical of Diffusion-RL prevent conventional batch-axis data parallelism (DP)
from forming sufficiently fine-grained pipeline units.

\textbf{Second, stage imbalance creates pipeline bubbles.}
Efficient overlap requires the generation and training times of each
micro-batch to be comparable. Although different numbers of GPUs can be
assigned to the two stages, choosing a balanced partition is difficult in
practice. Diffusion-RL hyperparameters, such as the fraction of timesteps
selected for training, affect both generation and training time. On a
heterogeneous cluster, runtime also varies with GPU compute capability,
memory capacity and bandwidth, and interconnect bandwidth and latency.
These observations motivate exploiting the computational structure of
diffusion inference and training while adaptively reusing trainer-side
bubbles for generation.

\textbf{Third, strict synchronization creates a final bubble.}
Synchronous Diffusion-RL inherently incurs an idle interval at the final
micro-batch of each global step. This bubble cannot be removed by ordinary
intra-step pipelining alone. It motivates overlapping bounded next-step
generation with the remaining synchronization while preserving trajectory
consistency.

Together, these observations motivate the fine-grained parallelism,
trainer-assisted generation, bubble-triggered scheduling, and
trajectory-consistent synchronization mechanisms introduced in the
following sections.

% !TEX root = ../main.tex
\section{System Overview of \sys{}}
\label{sec:system-model}

\sys{} follows a progressive optimization principle: after decoupling
generation and training, each mechanism targets the new bottleneck exposed by
the preceding mechanism. This sequence progressively improves utilization of
both resource pools and drives end-to-end performance toward the attainable
limit.

As shown in Figure~\ref{fig:4methods_com}(b), naive disaggregation first
decouples the resources used for generation and training, but exposes
mutual-waiting bubbles arising from the strong dependencies between the two
stages. Generation-axis pipelining (GAP) then creates finer-grained pipeline
units, allowing generation and training to overlap, as shown in
Figure~\ref{fig:4methods_com}(c). However, this finer pipeline granularity
reveals another bottleneck: small training micro-batches underutilize trainer
GPUs. Timestep parallelism (TSP) addresses this bottleneck by batching selected
denoising timesteps and recovering trainer-side compute efficiency, as shown in 
Figure~\ref{fig:4methods_com}(d).

As TSP shortens the processing time of each training micro-batch, the trainer
may finish before the generator produces the next input. Trainer-assisted
generation (TAG) therefore reuses otherwise idle trainer resources to execute
pending generation work, as illustrated in
Figure~\ref{fig:4methods_com}(e). TAG largely eliminates trainer-side idle
time under strict on-policy execution, but a final generator bubble remains at
the global-step boundary: the generator must wait for the trainer's latest
weights before starting the next rollout. Finally, trajectory-consistent stale
synchronization (TCSS) enables bounded asynchronous execution while
ensuring that each denoising trajectory uses a single policy snapshot. TCSS further improves generator
utilization, as shown in Figure~\ref{fig:4methods_com}(f).

% !TEX root = ../main.tex
\section{Diffusion-based Parallelism}
\label{sec:diffusion-parallelism}

\begin{figure}[t]
  \centering
  \includegraphics[width=\columnwidth]{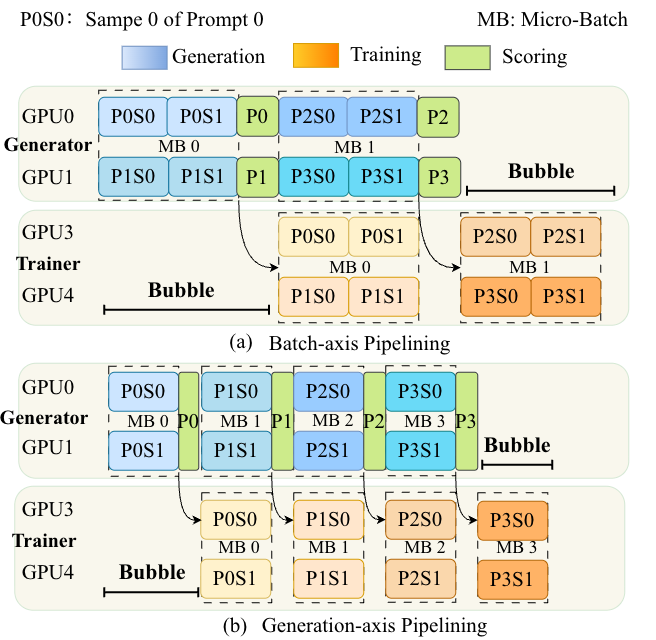}
  \caption{Comparison of micro-batch partitioning strategies. (a) Conventional
  partitioning along the batch dimension. (b) GAP partitions work along the
  generation axis, creating finer-grained pipeline units and enabling more
  overlap.}
  \label{fig:microbatch}
\end{figure}

\subsection{Generation-Axis Pipelining (GAP)}
\label{sec:gap}

Existing autoregressive RL (AR-RL) systems primarily construct micro-batches
along the prompt or sequence dimension and balance them according to token
length. This approach is effective for AR workloads, whose RL training batches
often contain hundreds or thousands of samples. A Diffusion-RL training batch,
however, often contains only dozens or even tens of samples. Naively
partitioning along the batch dimension therefore produces too few pipeline
units under small-batch Diffusion-RL settings, as shown in
Figure~\ref{fig:microbatch}(a).

\sys{} introduces generation-axis pipelining (GAP), a micro-batch formation
strategy that partitions work along the generation axis. GAP creates
finer-grained pipeline units while preserving the semantics of group reward
normalization and policy updates. Figure~\ref{fig:microbatch}(a) illustrates
the conventional batch-axis strategy. Different GPUs sample different prompts,
and the reward and advantage for a prompt can be computed only after all of its
samples finish generation. In the example, two GPUs process a training batch
of four samples, with two samples generated for each prompt. This configuration
can form at most two pipeline micro-batches.

GAP instead partitions work along the sampling dimension, as illustrated in
Figure~\ref{fig:microbatch}(b). Different GPUs sample the same prompt, and once
all GPUs complete the corresponding samples, their rewards and advantages can
be computed and passed to the trainer. Under the same configuration, GAP forms
four micro-batches. The additional micro-batches expose finer-grained
generation-training overlap without requiring a larger RL batch.

\textbf{Equivalence proof.} GAP does not change the
optimization objective. Let $R$ be the total number of generator workers. When $R$ divides
$M$, let $Q=M/R$. The GAP layout processes each prompt through $Q$ generation
rounds. In round $q$, rank $r$ produces generation $m=(q-1)R+r$,
yielding
\begin{equation}
  \label{eq:gap-gradient-equivalence}
  \begin{aligned}
    \nabla_\theta\mathcal{L}_{\mathrm{pg}}
      &=-\frac{1}{Z}\sum_{b=1}^{B}\sum_{q=1}^{Q}\sum_{r=1}^{R}
        \sum_{t\in\mathcal{S}_{b,m}}
        \nabla_\theta\mathcal{L}_{b,(q-1)R+r,t},
  \end{aligned}
\end{equation}
which is exactly the same sum as Eq.\ref{eq:normalized-policy-gradient} under the index mapping $m=(q-1)R+r$. When $M < R$ and $R = PM$, \sys{}
instead partitions the workers into $P$ groups of $M$ ranks. Ranks within each worker group generate
different samples for the same prompt, while different worker groups process different prompts in
parallel. 

\subsection{Timestep Parallelism (TSP)}
\label{sec:tsp}
However, this fine-grained pipeline exposes a new problem: the generator
forwards each result to the trainer as soon as it becomes available. Although
this enables timely training, each resulting micro-batch is small, preventing
the trainer from improving GPU utilization by increasing the micro-batch size.
To address this problem, \sys{} builds on GAP and employs timestep parallelism
(TSP). Eq.~\ref{eq:normalized-policy-gradient} shows that gradients from
individual timesteps are accumulated by summation. Whereas Diffusion-RL
algorithms such as DanceGRPO and FlowGRPO typically compute these timestep
gradients sequentially, \sys{} groups multiple timesteps into a timestep batch
and processes them in parallel, thereby maintaining high trainer utilization
even with small micro-batches.
This consideration becomes even more critical in sharded
training setups such as FSDP~\cite{zhao2023pytorch} and ZeRO~\cite{zero20}, in which parameters are
partitioned across devices and must be gathered before each layer is computed.
If selected timesteps are processed strictly one after another, each parameter
gather benefits only one timestep. By batching several timesteps, TSP reuses
each gathered parameter view for more computation, amortizing communication
cost and improving hardware utilization.

Together, GAP and TSP
improve both pipeline granularity and per-micro-batch compute efficiency.

% !TEX root = ../main.tex
\section{Elastic Trainer-Assisted Generation (TAG)}
\label{sec:tag}

\begin{figure}[t]
  \centering
  \includegraphics[width=1.0\columnwidth]{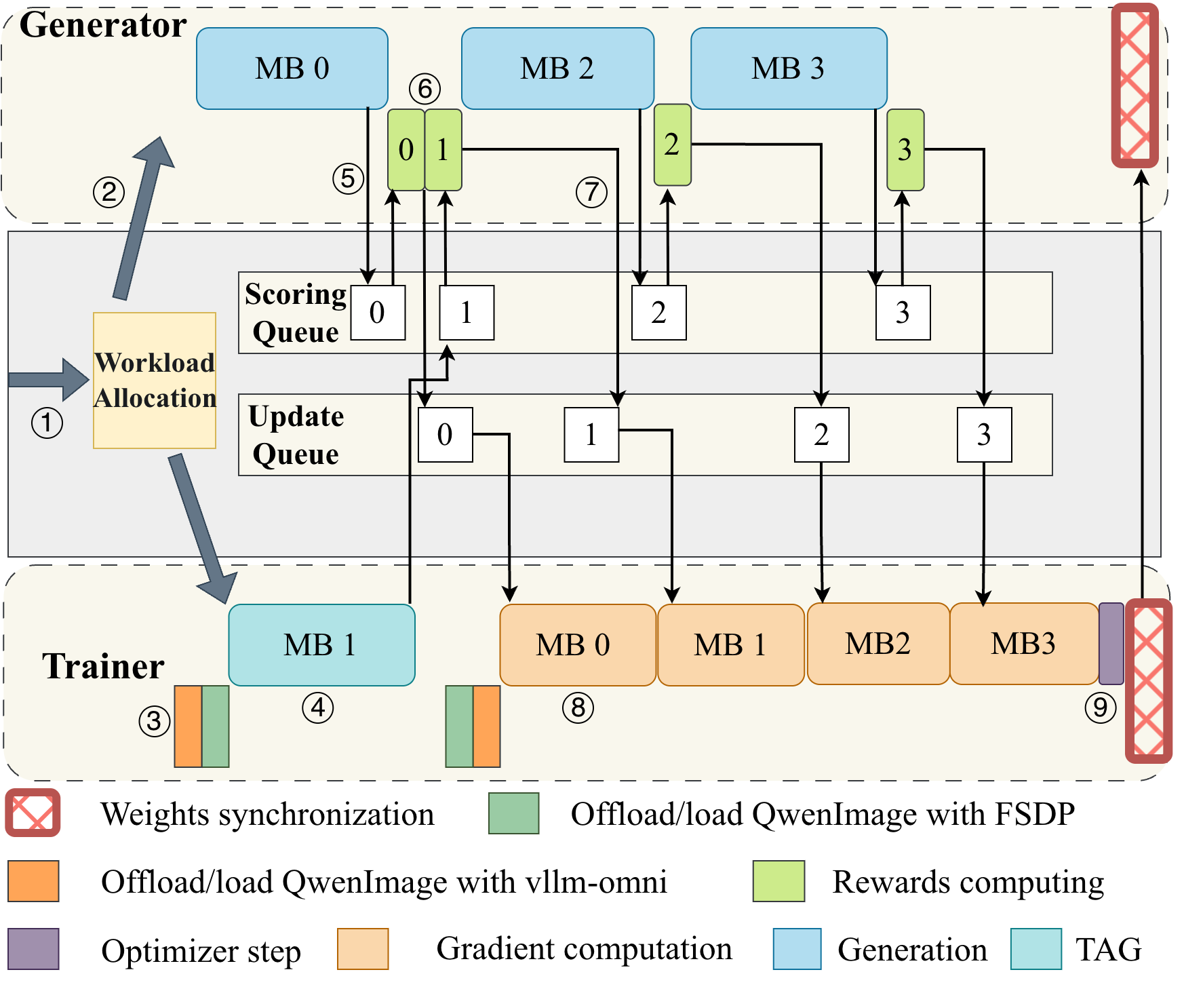}
  \caption{TAG execution workflow.}
  \label{TAG}
\end{figure}

\begin{figure}[t]
  \centering
  \includegraphics[width=1.0\columnwidth]{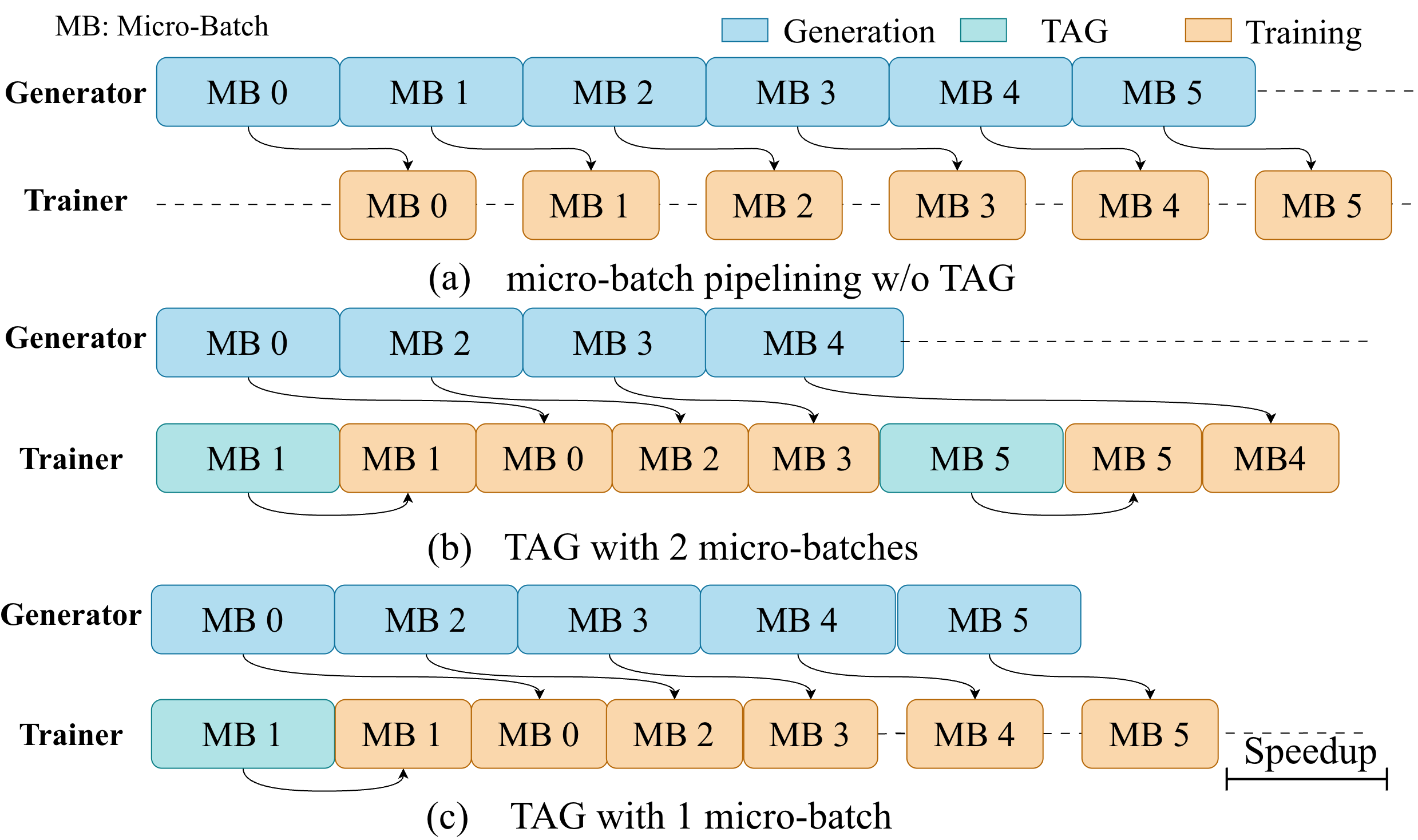}
  \caption{Impact of different numbers of TAG micro-batches on end-to-end time.}
  \label{TAG12}
\end{figure}

After addressing low trainer utilization under fine-grained pipelining, a new
bottleneck is exposed. As shown in Fig.~\ref{fig:4methods_com}(c)--(d),
TSP shortens the trainer-side processing time of each micro-batch, allowing the
trainer to finish a micro-batch faster than the generator. Generation therefore
becomes the new bottleneck. To address this imbalance, \sys{} adopts a
workload-balanced resource allocation strategy for
Diffusion RL, while allowing the trainer to take over part of the generation workload
when it would otherwise be idle. This mechanism is called \emph{Elastic Trainer-Assisted
Generation (TAG)}, as illustrated in Figure~\ref{TAG12}.

\subsection{TAG Execution Workflow}
\label{sec:tag-execution-workflow}
A practical RL system must decide how to construct resource pools, and which
components should be disaggregated or colocated. Taking VeRL-Omni as an example,
the system mainly consists of a generator and a reward model backed by
vLLM-Omni, and a trainer backed by FSDP. After separating the generator from
the trainer, \sys{} colocates the reward model with the generator in the same
resource pool, because the trainer has high memory demand during training and
is better suited to occupying an exclusive resource pool.

However, colocating scoring with generation under pipelined execution introduces
another issue: dynamically arriving tasks may compete for the same compute
resource. For example, after the trainer finishes a TAG task, the system may
request scoring for that generated result. If the generator is still executing
another generation task, the scoring task and generation task would contend for
the same resource pool, sharply reducing efficiency. 
\sys{} therefore uses a buffering and synchronization mechanism to coordinate
task execution. The driver maintains two queues: the Scoring Queue (SQ) buffers
generated images, videos, and intermediate states awaiting reward evaluation,
whereas the Update Queue (UQ) buffers reward-annotated samples awaiting a
trainer-side update. The circled numbers in Fig.~\ref{TAG} correspond to the
following execution steps.

\noindent\emph{Step~1 (Micro-batch allocation).} The driver performs centralized data
preprocessing and assigns pending micro-batches to the generator and trainer.
\par\noindent
\emph{Step~2 (Generation).} The generator starts normal rollout generation for
its assigned micro-batches.
\par\noindent
\emph{Step~3 (Model context switch).} Before the
trainer performs TAG, it switches model contexts when the generation and
training stages use different backends: it loads the generation model and
offloads the training model to release memory.
\par\noindent
\emph{Step~4 (TAG execution).}
When the trainer would otherwise remain idle, it generates an assigned TAG
micro-batch using the loaded generation model.

\noindent
\emph{Step~5 (Enqueue generated results).} Once either normal generation or TAG
finishes, the resulting images, videos, and intermediate states are pushed into
SQ.
\par\noindent
\emph{Step~6 (Reward computation).} After completing its current task, the
generator checks SQ. If SQ is non-empty, it prioritizes scoring the queued
results over starting another generation task and computes their rewards.
\par\noindent
\emph{Step~7 (Enqueue scored samples).} The reward-annotated samples are pushed
from SQ into UQ, making them available to the trainer as early as possible.

\noindent
\emph{Step~8 (Trainer update).} After TAG completes, the trainer restores the
training model when a context switch is required and checks UQ. If UQ is
non-empty, it consumes the ready samples and performs their trainer-side
updates; otherwise, the driver may assign another TAG task according to the TAG
scheduling policy. Each update micro-batch only accumulates gradients and does
not immediately modify the policy weights.
\par\noindent
\emph{Step~9 (Optimizer step and
weight synchronization).} After all update micro-batches in the current global
step have been trained, the trainer invokes the optimizer step and synchronizes
the updated policy weights with the generator. Consequently, the overall
workflow preserves on-policy semantics.

This workflow avoids severe efficiency degradation caused by
resource contention, while also preventing the trainer from waiting idly for new training samples. A remaining question is when TAG should be applied. \sys{}'s search strategy is described in Section~\ref{sec:tag-search}.

\subsection{Recursive Bubble-Triggered TAG Search}
\label{sec:tag-search}

\begin{figure}[t]
  \centering
  \includegraphics[width=\columnwidth]{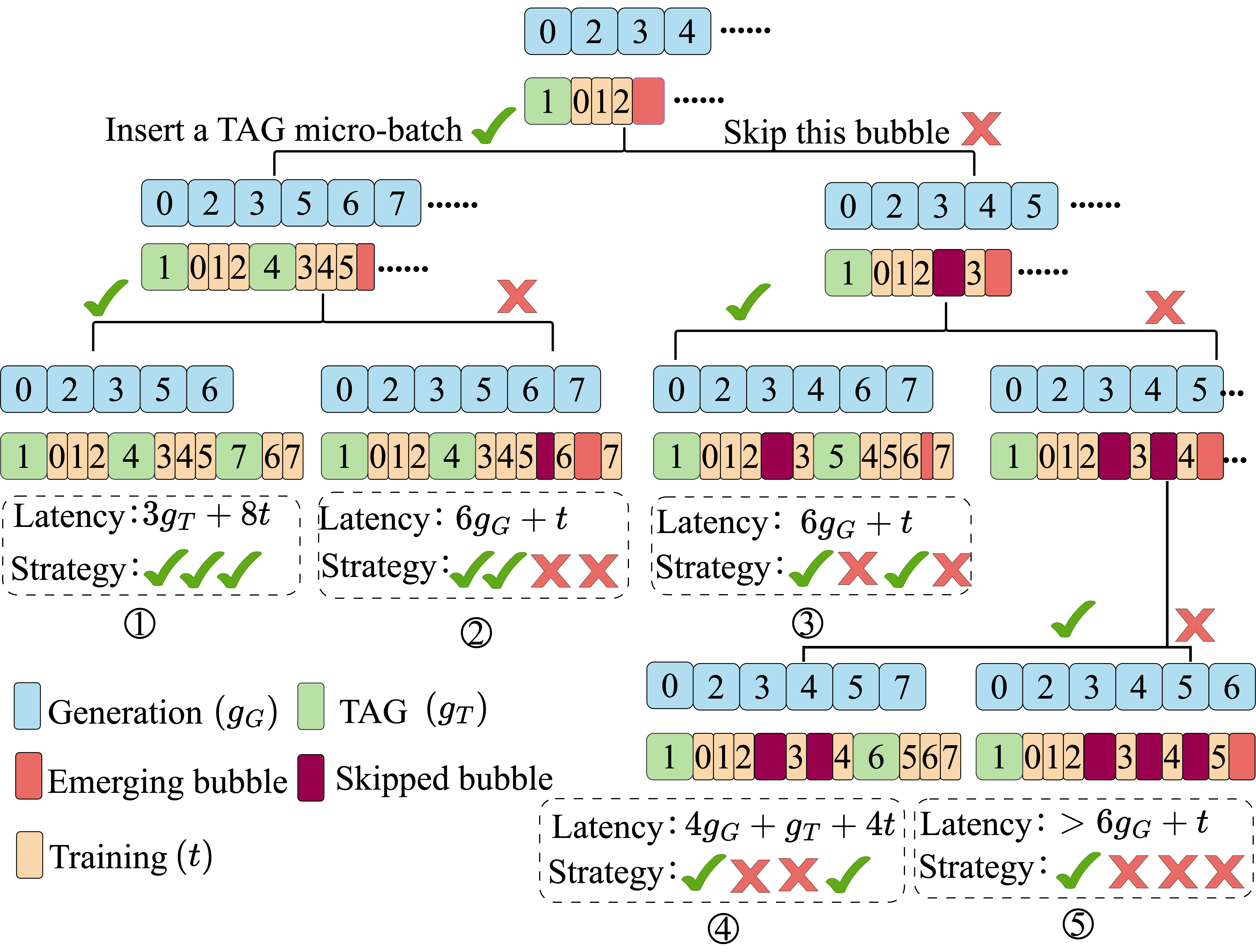}
  \caption{Recursive bubble-triggered TAG search. At each emerging bubble,
  the scheduler either inserts one TAG micro-batch or skips the bubble.}
  \label{fig:search}
\end{figure}

\sys{} uses a recursive global search strategy,
called \textbf{Bubble-Triggered TAG Search} (BTS), to decide whether each emerging bubble
should be filled by TAG. Because inserting a TAG
micro-batch can fill the current trainer-side bubble, but it may also delay
the training of subsequent micro-batches as shown in
Figure~\ref{TAG12}; in other words, blindly inserting TAG
at every bubble can be locally beneficial but globally suboptimal.

Figure~\ref{fig:search} illustrates the decision tree. When a new bubble emerges, the scheduler decides whether to fill it
with one TAG micro-batch or skip it. A skipped bubble is marked inactive and will
not be reconsidered. The recursion terminates in two cases. First, even if a
bubble appears, there may be no remaining micro-batch that can be assigned to
TAG, as illustrated by nodes 2 and 3 in Figure~\ref{fig:search}. Second, with
branch-and-bound pruning, a branch can be stopped early once its latency lower
bound already exceeds the current best latency, as illustrated by node 5.

In practical training, \sys{} applies TAG to the first bubble by default, i.e., the
position of micro-batch 1 in Figure~\ref{fig:search}. This provides online
profiling measurements for the generation time of the next micro-batch on the
generator, as well as the generation and training times on the trainer. Training
then proceeds normally. Although Diffusion-RL does not exhibit the same long-tail
effect as AR workloads and its micro-batch runtimes are usually similar, small
variations can still arise from hardware variability, reward computation, and
cold-start effects during online profiling. These measurement errors may accumulate as execution
progresses, so a one-time global search is insufficient. Instead, whenever a
bubble is encountered, \sys{} updates the runtime estimates using historical online
profiling measurements, uses the averaged values as the new search input, and
recomputes the strategy to correct drift caused by measurement error.

\begin{figure}[t]
  \centering
  \includegraphics[width=\columnwidth]{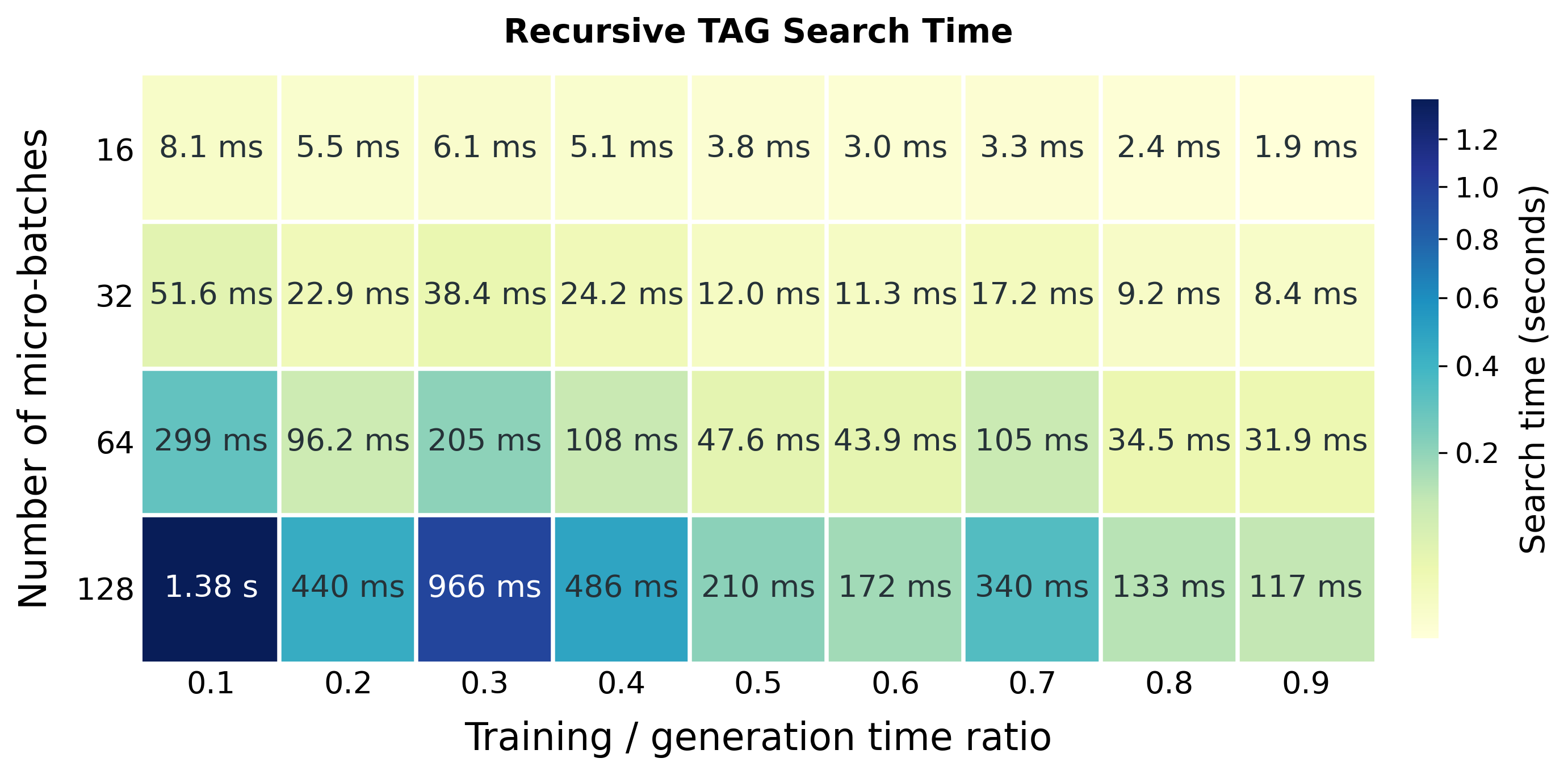}
  \caption{Recursive TAG search time.}
  \label{searchcost}
\end{figure}

If the execution exposes $K$ bubbles, BTS
has worst-case complexity $O(2^K)$, or $O(B2^K)$ when each state simulation
costs $O(B)$ for $B$ micro-batches. However, the homogeneous micro-batch model
allows \sys{} to substantially reduce the effective search space. Since all
micro-batches have the same generation, TAG, and update costs, their identities
are irrelevant to future scheduling decisions. \sys{} therefore compresses a schedule
state into aggregate counters, including the numbers of pending, ready-to-update,
and completed micro-batches, together with the current generator and trainer
timesteps. Different decision paths that reach the same
compressed state share the same optimal continuation. \sys{} memoizes the optimal
result for each compressed state and reuses it when the state is encountered
again. For example, nodes 2 and 3 in Figure~\ref{fig:search} follow different
decision paths, but they reach equivalent compressed states and therefore have
the same remaining optimal latency.
Figure~\ref{searchcost} reports the runtime of a single search. The y-axis
varies the number of micro-batches, where a larger value implies a larger
search space. The x-axis varies the ratio between the update time and the
generation time of one micro-batch, where a smaller ratio indicates more
frequent pipeline bubbles. Even in the most challenging setting
$(B=128,\; t/g_G=0.1)$, the search overhead remains negligible compared with
generation and training.

% !TEX root = ../main.tex
\section{Trajectory-Consistent Stale Synchronization (TCSS)}
\label{sec:tcss}

As shown in Figure~\ref{fig:4methods_com}(f), the generator idle time during the
final trainer-side micro-batch update is unavoidable under strict
synchronization. The generator cannot start producing the next rollout batch until
the latest trainer weights have been synchronized. This barrier remains even
after micro-batch pipelining and TAG, because it is caused by the global-step
consistency requirement rather than by coarse scheduling granularity.

To overlap this bubble, \sys{} employs \emph{Trajectory-Consistent Stale
Synchronization} (TCSS). Let $\pi_k$ be the policy available to the generator
at the boundary between global steps $k$ and $k+1$. While the trainer
synchronizes the updated snapshot $\pi_{k+1}$, TCSS may start a bounded subset
of step-$k+1$ rollouts using $\pi_k$; rollouts started after synchronization use
$\pi_{k+1}$.

Let $v_b^{(k+1)}$ be the behavior-policy version assigned to prompt-group
micro-batch $b$, which contains all $M$ trajectories generated for prompt $b$.
TCSS defines the stale fraction as
\begin{equation}
  \begin{aligned}
    s_{k+1}
      &= \frac{1}{B}
        \sum_{b=1}^{B}
        \mathbf{1}\!\left[v_b^{(k+1)}=k\right]
        \le s_{\max},\\
    v_b^{(k+1)}
      &\in\{k,k+1\}, \qquad \forall b.
  \end{aligned}
  \label{eq:tcss-stale-fraction}
\end{equation}
where the numerator counts prompt-group micro-batches generated by $\pi_k$, the
denominator is the total number of prompt-group micro-batches in step $k+1$,
and $s_{\max}$ is the staleness threshold. The version constraint permits
$\pi_k$ but excludes $\pi_{k-1}$ and older snapshots. Hence, ``one-step stale''
refers to policy-version distance, not to one stale micro-batch. The scheduler
enforces this constraint with a budget of at most $s_{\max}B$ stale
micro-batches.

\begin{figure}[t]
  \centering
  \includegraphics[width=\columnwidth]{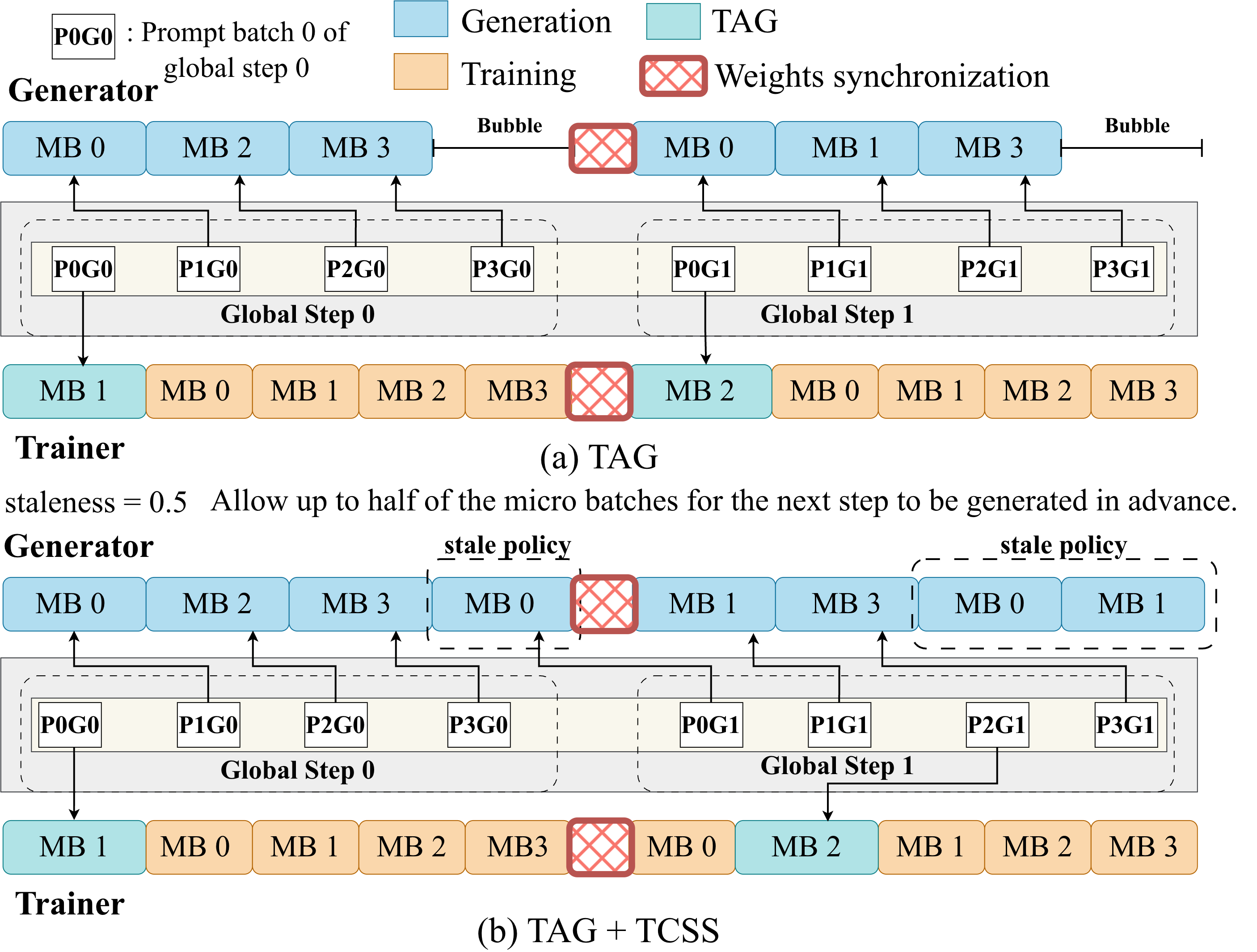}
  \caption{Comparison between TAG and TCSS.}
  \label{TCSS}
\end{figure}

TCSS allows versions to differ across prompt-group micro-batches, but all $M$
trajectories in micro-batch $b$ share $v_b^{(k+1)}$. All denoising transitions
of each trajectory $\tau_{b,m}^{(k+1)}$ therefore satisfy
\begin{equation}
  x_{b,m,t-1}^{(k+1)}
    \sim \pi_{v_b^{(k+1)}}
      \!\left(\cdot\mid x_{b,m,t}^{(k+1)},c_b\right).
  \label{eq:tcss-trajectory-consistency}
\end{equation}
Accordingly, every $\ell^{\mathrm{old}}_{b,m,t}$ in that trajectory is computed
under the same snapshot $\pi_{v_b^{(k+1)}}$. Equations~\ref{eq:importance-ratio}
and~\ref{eq:clipped-policy-objective} thus use the matching behavior-policy
log-probability rather than a mixture of versions. Unlike asynchronous AR
schemes that may partition a response along token prefixes, TCSS mixes versions
only across complete prompt-group micro-batches. Figure~\ref{TCSS} shows how
this constraint overlaps part of the global-step boundary bubble.

Stale trajectories still constitute bounded off-policy data; clipping in
Eq.~\ref{eq:clipped-policy-objective} does not eliminate this bias. Therefore,
$s_{\max}$ is a system--algorithm tradeoff that must be chosen conservatively.
Section~\ref{sec:evaluation} evaluates reward behavior under the selected
threshold.

% !TEX root = ../main.tex
\section{Evaluation}
\label{sec:evaluation}

\subsection{Experimental Settings}
\paragraph{Testbed.} Experiments are mainly carried out on:
1) Testbed A, a 32-GPU cluster comprising four interconnected 8-GPU nodes over a 200 Gb/s network, with each 48 GB GPU providing 336 third-generation Tensor Cores and up to 38.7 FP32 TFLOPS / 309.7 Tensor TFLOPS peak performance.
2) Testbed B, a 32-GPU cluster comprising four interconnected 8-GPU nodes over a 400 Gb/s network, with each 80 GB GPU providing 528 fourth-generation Tensor Cores and up to 67 FP32 TFLOPS / 1,979 BF16/FP16 Tensor TFLOPS peak performance.
3) Testbed C, a 16-GPU cluster comprising two interconnected 8-GPU nodes over a 200 Gb/s network, with each 96 GB GPU providing 752 fifth-generation Tensor Cores and up to 126 FP32 TFLOPS / 503.8 BF16/FP16 Tensor Core TFLOPS peak performance.

\begin{figure}[t]
  \centering
  \includegraphics[width=1.0\columnwidth]{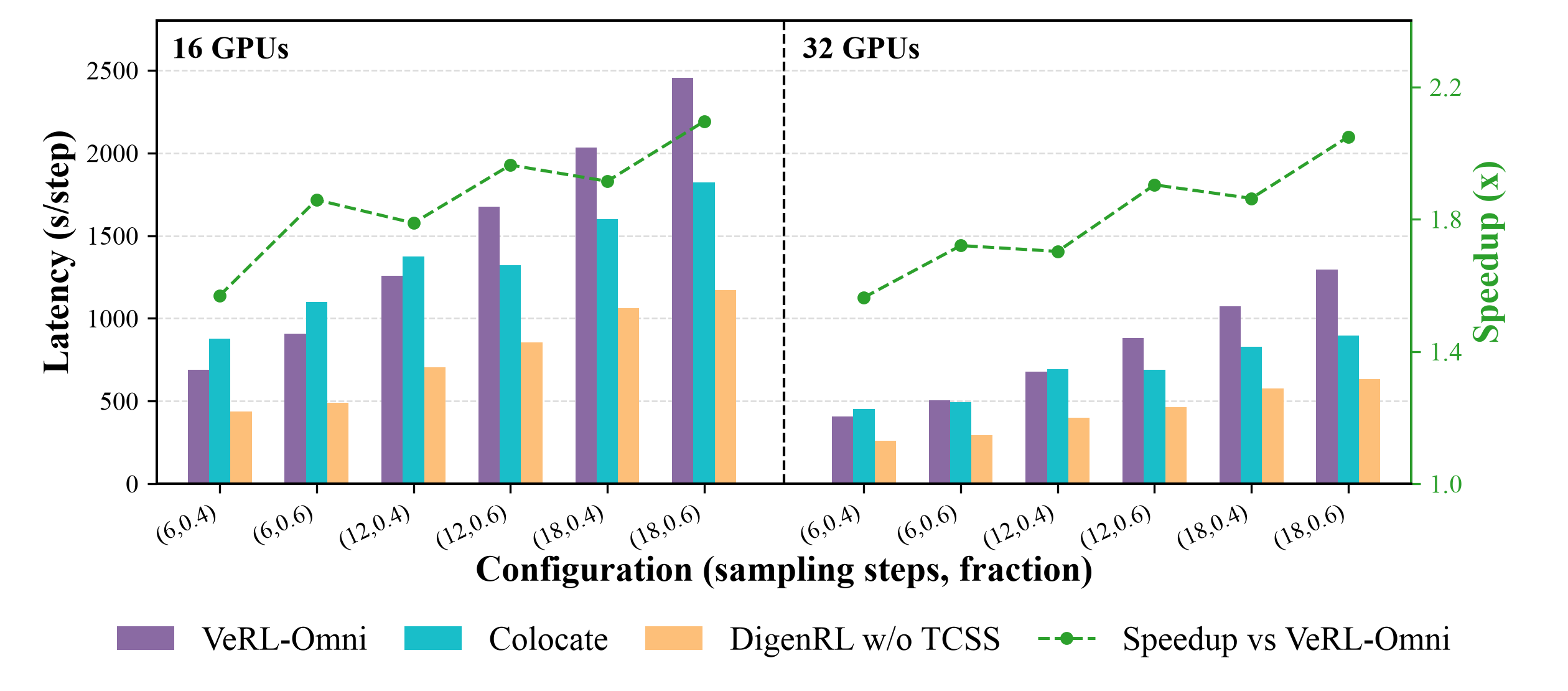}
  \caption{Qwen-Image end-to-end time comparison with VeRL-Omni, colocated execution and DigenRL.}
  \label{verl_omni_colocate_tag}
\end{figure}
\begin{figure}[t]
  \centering
  \includegraphics[width=1.0\columnwidth]{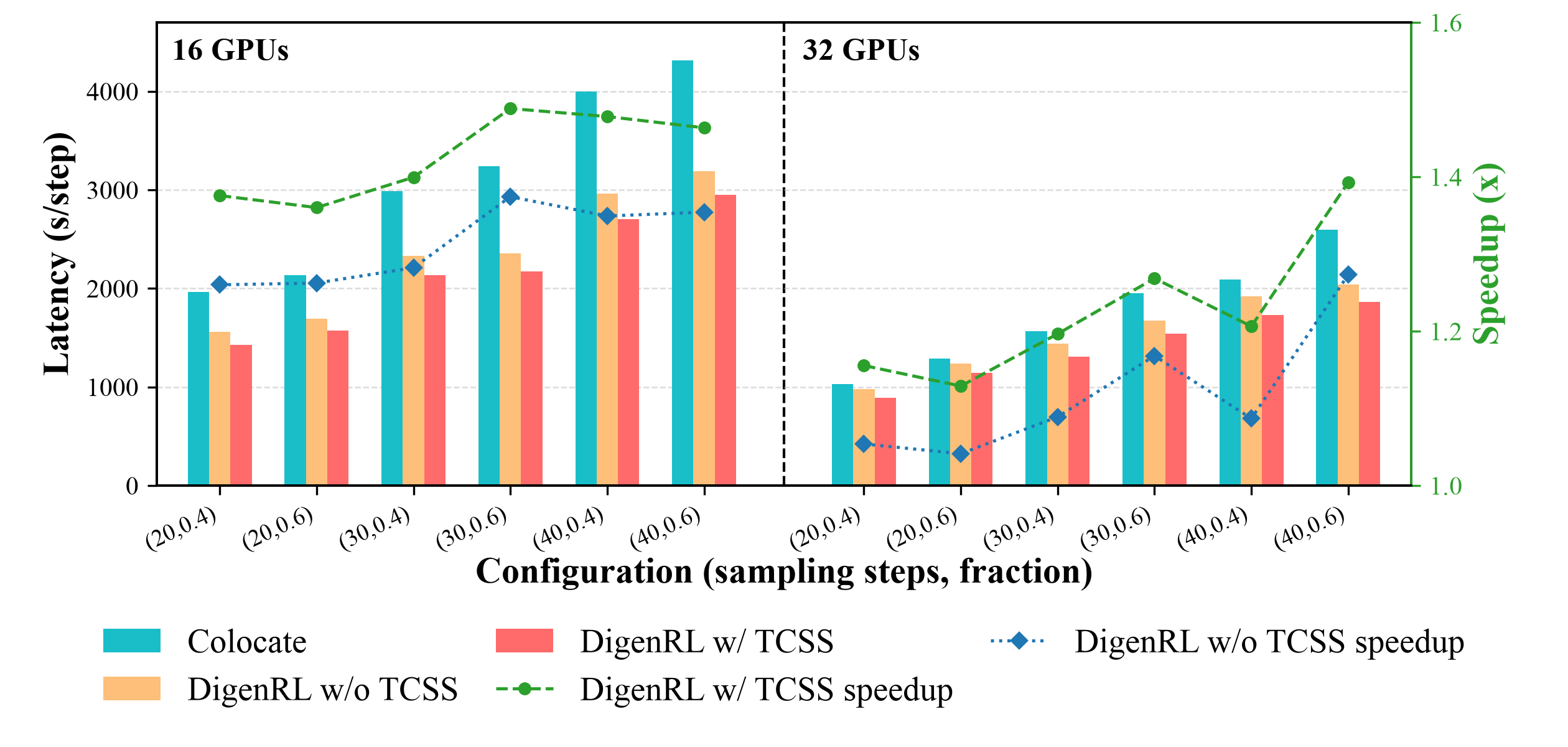}
  \caption{HunyuanVideo end-to-end time comparison with colocated execution and DigenRL.}
  \label{hunyuanvideo_colocate_tag_tcs}
\end{figure}
\paragraph{Baselines.}
Our end-to-end evaluation consists of four parts. 
First, to validate the effectiveness of DigenRL, we integrate the Diffusion RL algorithm FlowGRPO and support the QwenImage 20B model. We compare its end-to-end time against VeRL-Omni, a state-of-the-art open-source Diffusion RL framework. For fairness, DigenRL uses the same inference backend, vLLM-Omni, and training backend, FSDP, as VeRL-Omni.
Second, to verify that disaggregated execution outperforms colocated execution, we further integrate HunyuanVideo with DanceGRPO and Wan2.1 14B with FlowGRPO into DigenRL, and compare their end-to-end time against their native colocated execution.
Third, to evaluate DigenRL under heterogeneous resources, we compare it with the colocated execution of HunyuanVideo and show that it can still maintain high throughput.
Finally, we conduct ablation experiments to break down the gains from each optimization. We also verify that, when bounding the number of stale samples and limiting their impact on training, TCSS maintains a reward curve comparable to on-policy training.

\subsection{End-to-End Time Comparison with Open-Source Frameworks}

Our experiments are conducted on Testbed B with a training batch size of 32 and an image resolution of 512 × 512. We use intra-node FSDP full sharding and inter-node data parallelism (DP). Since FlowGRPO for QwenImage uses Qwen3-VL-8B-Instruct~\cite{qwen3vl25} as the reward model, we colocate the generator and reward model to balance GPU memory pressure. Generation and scoring are then coordinated through the two-queue scheduling mechanism described in \S\ref{sec:tag}.

Across all cases, DigenRL substantially outperforms both native colocation and
VeRL-Omni. Compared with VeRL-Omni, DigenRL achieves speedups ranging from
1.56$\times$ to 2.10$\times$. A key reason is that vLLM-Omni schedules multiple
requests through asynchronous queuing: as the batch size grows, scheduling
queues and long-tail requests increasingly limit efficiency, and larger batches
can even reduce overall throughput. This effect appears in both VeRL-Omni and
native colocation. DigenRL mitigates the issue by splitting large batches into
smaller micro-batches and pipelining them while controlling concurrency.

\begin{figure}[t]
  \centering
  \includegraphics[width=1.0\columnwidth]{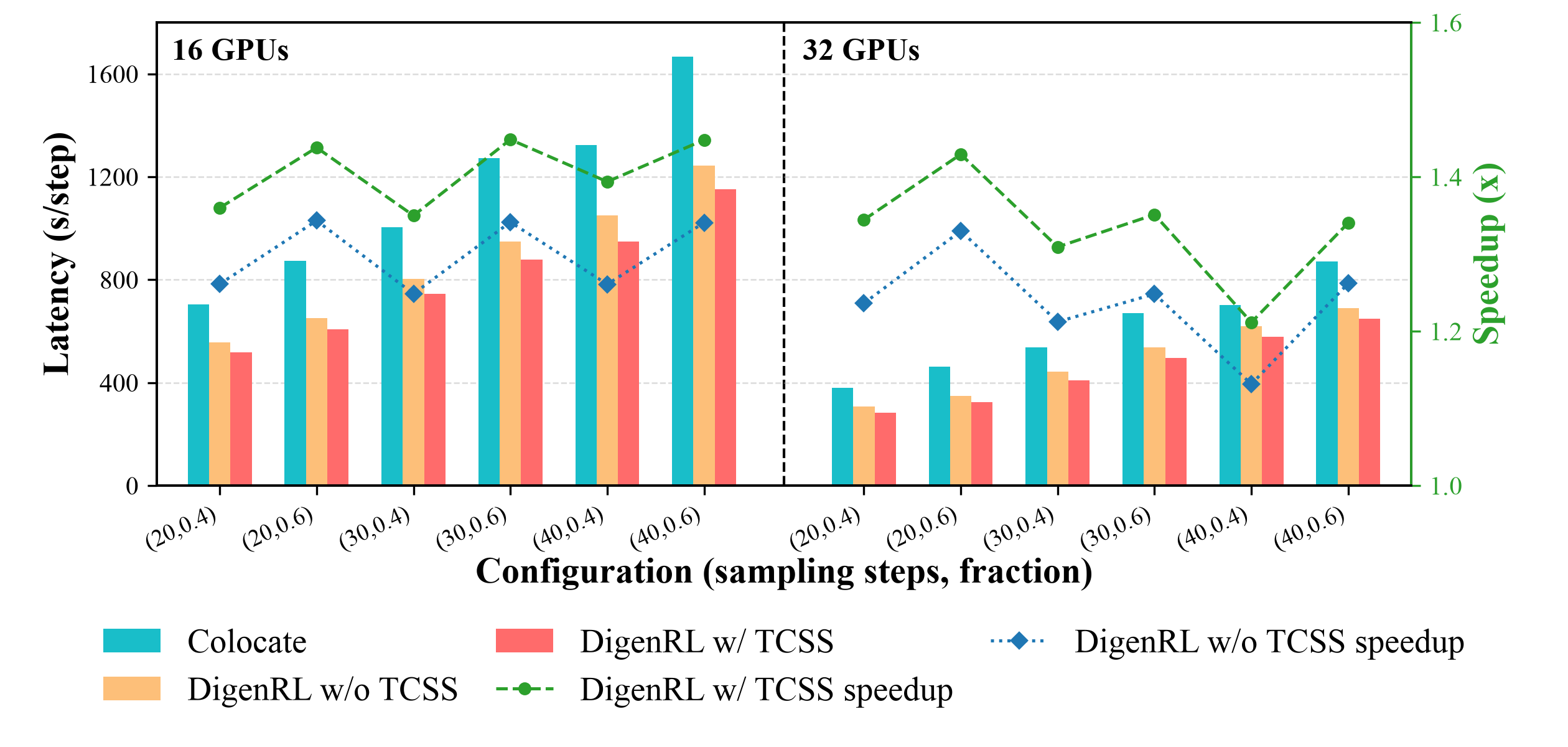}
  \caption{Wan2.1-14B end-to-end time comparison with colocated execution and DigenRL.}
  \label{wan14b_colocate_tag_tcss}
\end{figure}

\subsection{End-to-End Time Comparison with Native Colocation}\label{sec:colocate_compare}
Figure~\ref{hunyuanvideo_colocate_tag_tcs} and Figure~\ref{wan14b_colocate_tag_tcss} compare \sys{} with native
colocated Diffusion RL execution. The HunyuanVideo experiments are conducted on
Testbed A, and the Wan2.1 14B experiments are conducted on Testbed B. \sys{} splits
the nodes evenly between the generator and trainer. The model uses FSDP as the
backend, and the training batch size is 32. The evaluation uses common sampling-step counts of 20, 30,
and 40, and timestep fractions of 0.4 and 0.6, which
denotes the proportion of denoising timesteps used for training. The resolution
is $352 \times 352 \times 25$.

For HunyuanVideo, \sys{} achieves end-to-end speedups ranging from 1.04$\times$ to 1.37$\times$. When TCSS is enabled, the speedups increase to 1.13$\times$--1.49$\times$. The gains are larger on 16 GPUs because scaling to 32 GPUs introduces additional driver-side scheduling pressure and GPU--CPU communication overhead, which partially offset the benefits of these optimizations.

For Wan2.1 14B, \sys{} achieves end-to-end speedups ranging from 1.13$\times$ to 1.34$\times$. When TCSS is enabled, the speedups increase to 1.21$\times$--1.45$\times$. More importantly, it achieves similar gains on both 16 and 32 GPUs, mainly because Testbed B
provides stronger CPU scheduling capability and higher CPU-GPU communication
bandwidth than Testbed A, which mitigates the scheduling and communication pressure caused by scaling up. Another supporting observation
is that \sys{} gains increase with the number of sampling steps, because
computation occupies a larger fraction of the end-to-end time and therefore
partially amortizes communication and scheduling overhead.

\begin{figure}[t]
  \centering
  \includegraphics[width=1.0\columnwidth]{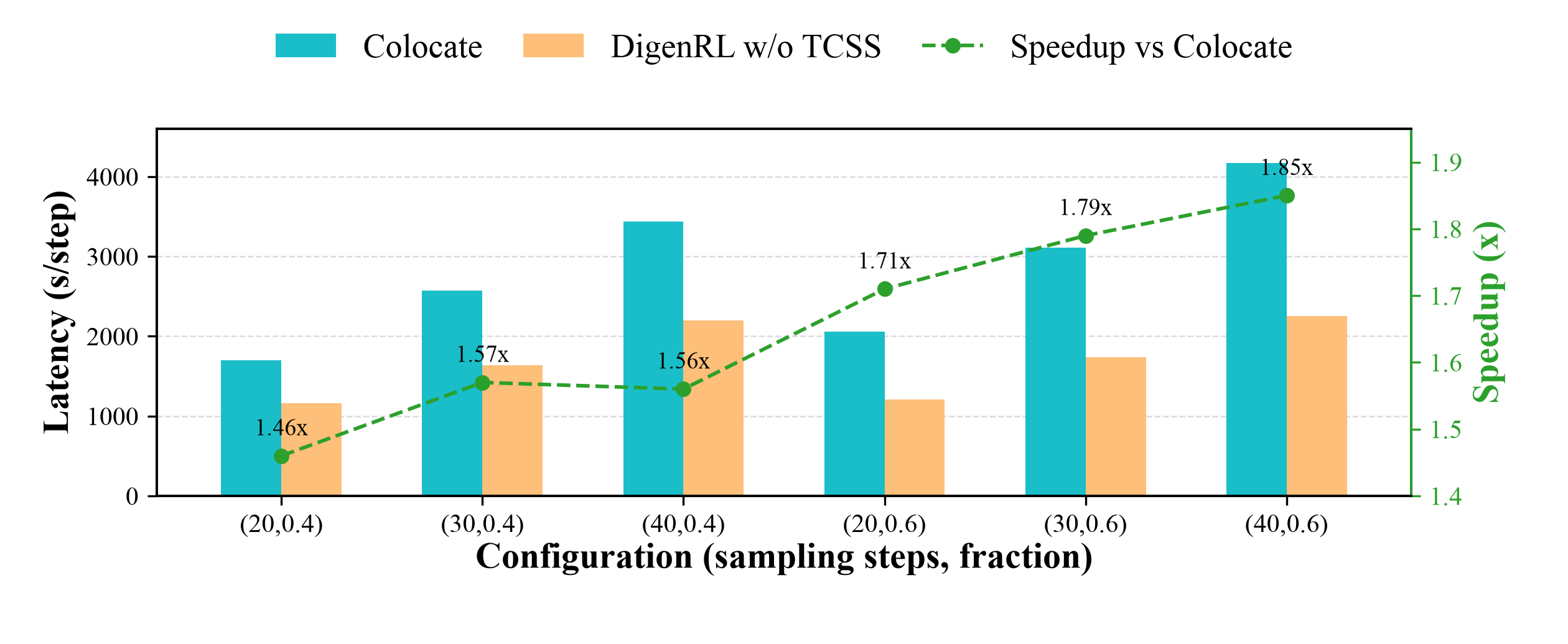}
  \caption{HunyuanVideo end-to-end latency on heterogeneous resources.}
  \label{hete}
\end{figure}

\subsection{End-to-End Time Comparison with Heterogeneous Resources}
These experiments use eight GPUs from Testbed A and eight
GPUs from Testbed C, connected through 200 Gb/s HDR InfiniBand. The evaluation uses
HunyuanVideo as the evaluation model because RL for video diffusion models
typically imposes higher compute and memory requirements, making the impact of
heterogeneous resources more pronounced. As shown in Figure~\ref{hete}, \sys{}
improves speedup over colocated execution from 1.46$\times$ to 1.85$\times$. The overall speedup increases with the number of sampling steps and the timestep fraction, as larger values of these parameters amplify performance differences across heterogeneous hardware.

Colocated execution is less effective under heterogeneous resources due to two key drawbacks. First, the more capable side often remains idle while waiting for the weaker side to complete the current stage, resulting in suboptimal resource utilization. Second, the training workload is constrained by the memory capacity of the weaker side, limiting support for higher-resolution videos and images. In contrast, \sys{} is well suited to this heterogeneous training setting. By assigning stronger compute resources to the trainer, it provides a higher memory ceiling for training, while allowing idle trainer resources to assist generation and thereby improving overall resource utilization.

\subsection{Training Effectiveness of TCSS}
\begin{figure}[!t]
  \centering
  \includegraphics[width=1.0\columnwidth]{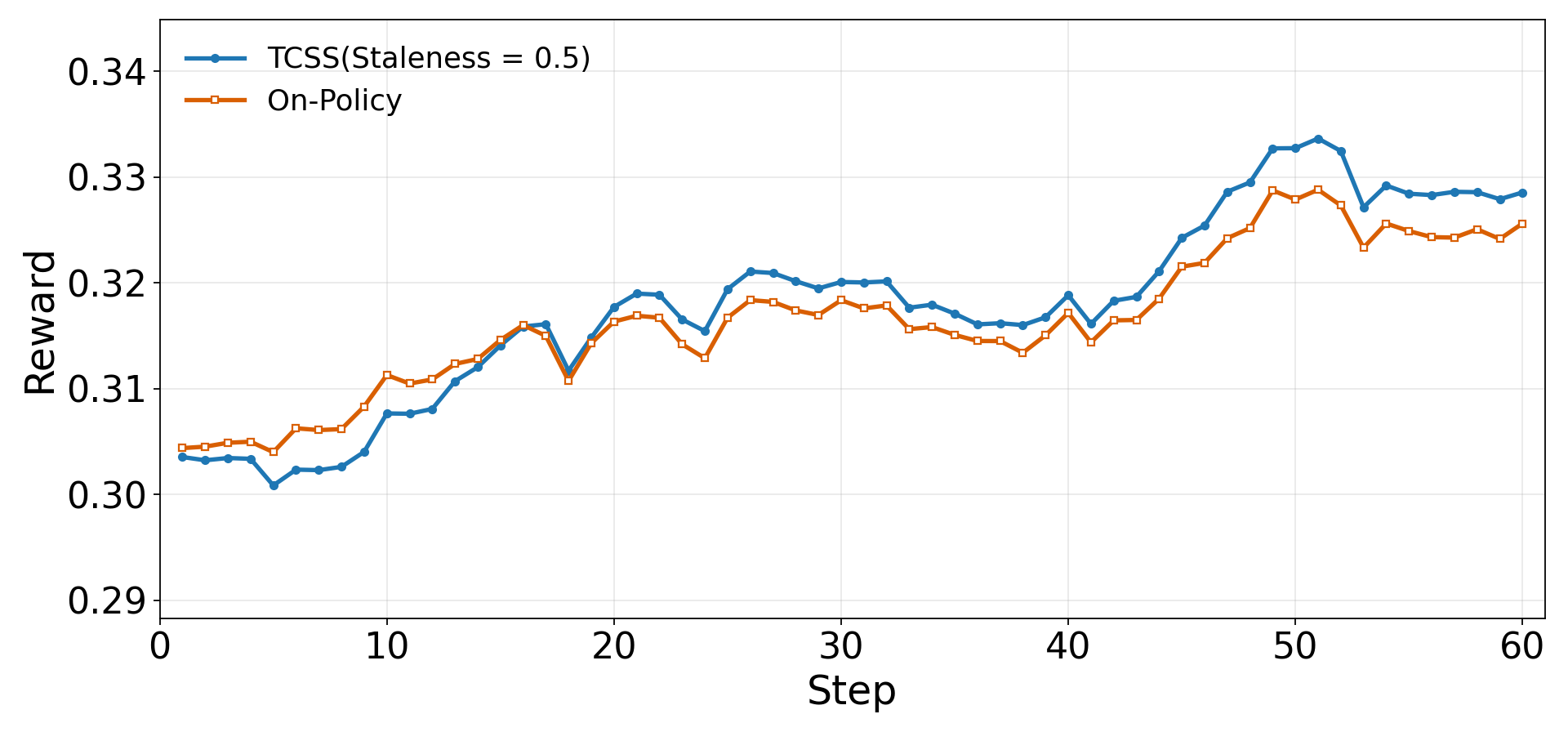}
  \caption{TCSS reward curve.}
  \label{TCSS_reward}
  \vspace{-0.8em}
\end{figure}

Figure~\ref{TCSS_reward} shows the reward curves of on-policy training and TCSS
when training FLUX.1-dev 12B. By strictly
controlling staleness and limiting the impact of stale policies during training,
TCSS maintains a steadily increasing reward even when the staleness threshold is
0.5.

\begin{figure}[t]
  \centering
  \includegraphics[width=1.0\columnwidth]{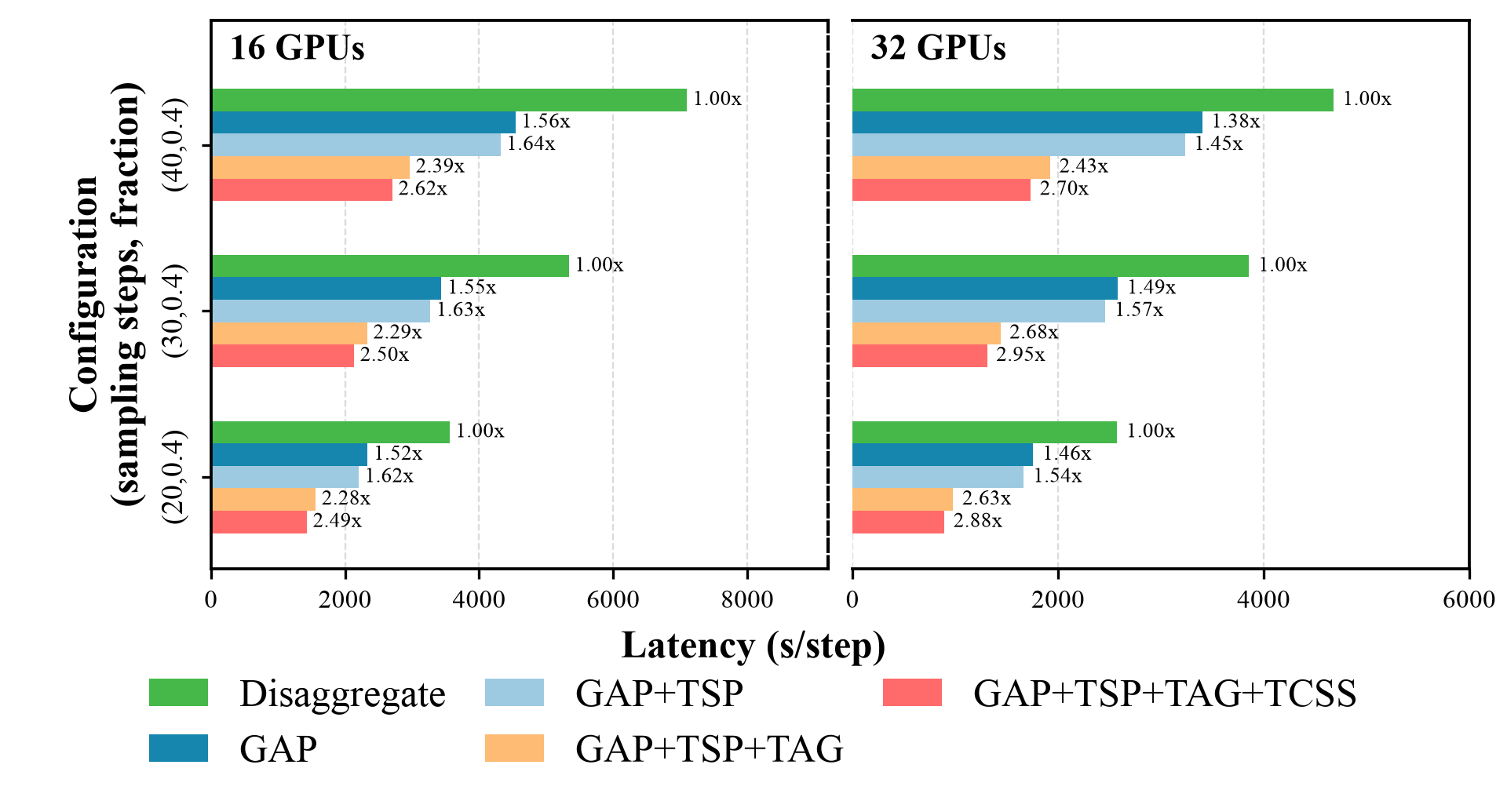}
  \caption{End-to-end latency of different optimization stages on HunyuanVideo. Labels show speedup relative to naive disaggregation.}
  \label{hunyuan_end}
\end{figure}
\begin{figure}[t]
  \centering
  \includegraphics[width=1.0\columnwidth]{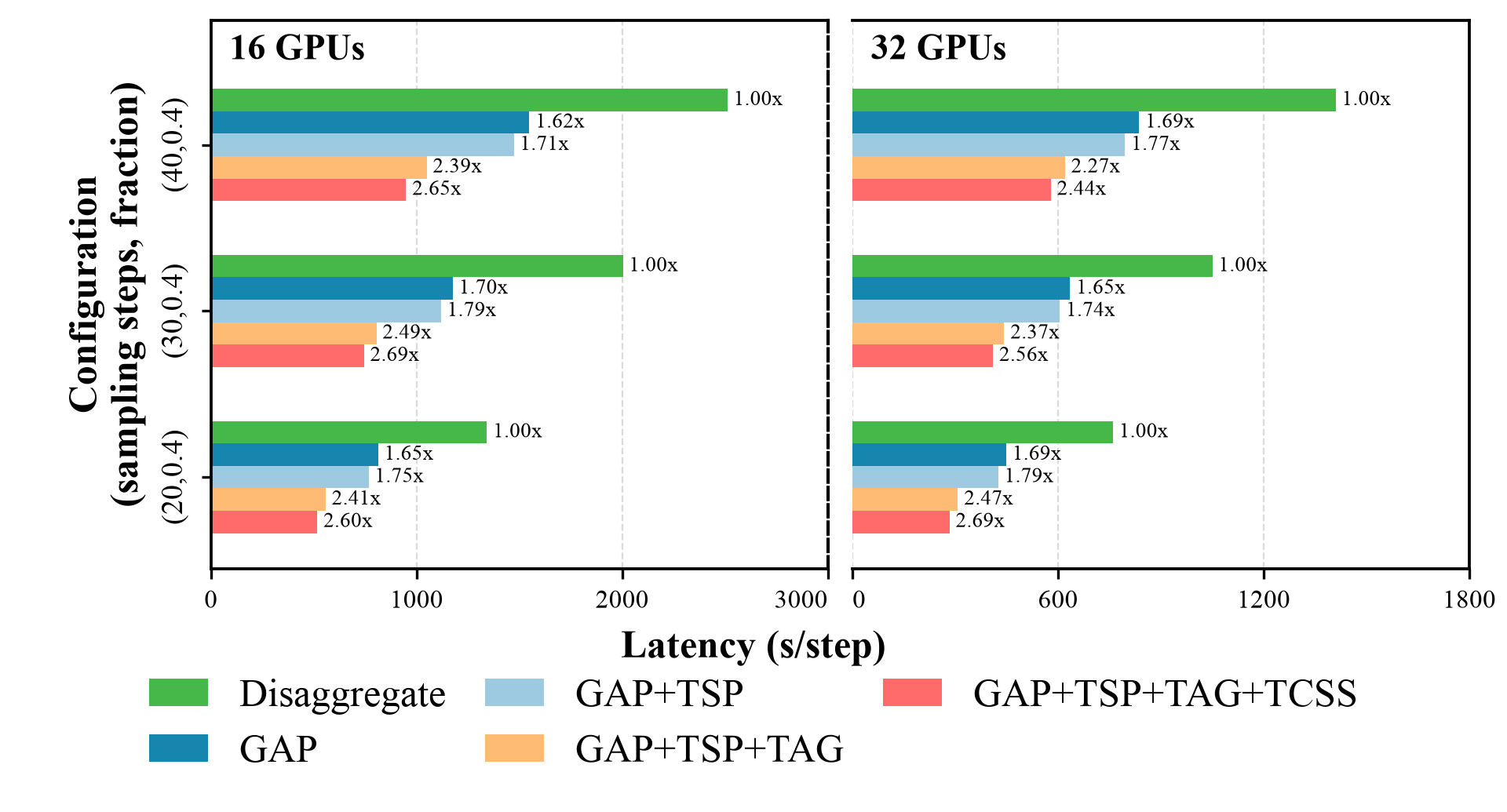}
  \caption{End-to-end latency of different optimization stages on Wan2.1-14B. Labels show speedup relative to naive disaggregation.}
  \label{wan_end}
\end{figure}
\subsection{Ablation Study}
Figure~\ref{hunyuan_end} shows the speedup brought by the optimizations introduced
on top of naive disaggregation for the 0.4 configurations. Across the three configurations and both cluster sizes, GAP provides 1.38--1.56$\times$ speedup. TSP reduces the training time of each micro-batch and provides 1.45--1.64$\times$ speedup. At this point, micro-batch training is no longer the bottleneck; generation becomes the dominant cost. TAG then alleviates this generation bottleneck by utilizing idle trainer resources, increasing the speedup to 2.28--2.68$\times$, while TCSS further increases it to 2.49--2.95$\times$. The additional gain
from TCSS is relatively small because
TAG with TSP already achieves high resource utilization; TCSS mainly extracts the
remaining bubbles at synchronization boundaries.

For Wan2.1-14B, as shown in Figure~\ref{wan_end}, GAP provides 1.62--1.70$\times$ speedup across the 0.4 configurations. TSP increases it to 1.71--1.79$\times$, TAG improves it to 2.27--2.49$\times$, and TCSS further increases it to 2.44--2.69$\times$.

% !TEX root = ../main.tex
\section{Related Work}
\label{sec:related-work}

\textbf{RL post-training systems.}
Recent RL post-training systems for autoregressive foundation models have made
large-scale policy optimization increasingly practical. General-purpose systems
such as veRL~\cite{hybridflow24}, OpenRLHF~\cite{openrlhf24}, ROLL~\cite{roll25}, and LlamaRL~\cite{llamarl25} provide modular execution
substrates for algorithms such as PPO~\cite{ppo17}, GRPO~\cite{mixgrpo25}, DPO~\cite{dpo23}, and ReMax~\cite{li2024remax}, and expose
distributed training backends for large models.
Disaggregated RL systems (AReaL~\cite{areal25}, StreamRL~\cite{streamrl25}, Laminar~\cite{sheng2026laminar}, RollArt~\cite{gao2026rollart}, Weave~\cite{wu2026weave}, OrchestrRL~\cite{tan2026orchestrrl}, etc.) for post-training text-only LLMs or multimodal LLMs have been proposed to improve the training throughput in different aspects. AReaL~\cite{areal25} eliminates the synchronization bottleneck of traditional RL pipelines through fully asynchronous training. StreamRL~\cite{streamrl25} further improves disaggregated RL execution by addressing pipeline and workload-skew bubbles. Laminar~\cite{sheng2026laminar} enhances resource utilization through asynchronous overlap between rollout and training. RollArt~\cite{gao2026rollart} focuses on scalable multi-task agentic RL training, while Weave~\cite{wu2026weave} improves cluster efficiency via co-scheduling of disaggregated RL workloads. OrchestrRL~\cite{tan2026orchestrrl} jointly optimizes compute and network resource orchestration for large-scale disaggregated RL deployments. RLBoost~\cite{rlboost26} exploits otherwise idle training
resources through partial response seeding. These systems
mainly target autoregressive decoding, where requests can often be partitioned
or resumed along token prefixes. \sys{} instead targets Diffusion-RL, where a
sample is generated through a tightly coupled denoising trajectory and training
must preserve step-level log-probabilities, rewards, and policy-version
consistency.

% !TEX root = ../main.tex
\section{Conclusion}
\label{sec:conclusion}

This paper presents \sys{}, a disaggregated RL system for post-training large
visual diffusion and flow models. \sys{} combines generation-axis pipelining
and timestep parallelism to improve pipeline granularity and trainer
utilization, TAG to reuse idle trainer resources for rollout generation, and
TCSS to overlap global-step synchronization while preserving trajectory-level
policy consistency. Experiments across four models and three 16--32-GPU
testbeds show that \sys{} improves end-to-end throughput over veRL-Omni and
native colocated execution. These results show that efficient disaggregation
requires coordinated optimization because removing one bottleneck shifts the
critical path.

\clearpage
\balance
\bibliographystyle{IEEEtran}
\bibliography{digenrl}

\end{document}